\documentclass{article}

\usepackage{microtype}
\usepackage{graphicx}
\usepackage{booktabs} 

\usepackage{hyperref}

\usepackage[accepted]{icml2020}

\usepackage{amsfonts}
\usepackage{subcaption}
\usepackage{amsmath}
\usepackage[ruled,vlined]{algorithm2e}

\makeatletter
\newcommand{\removelatexerror}{\let\@latex@error\@gobble}
\makeatother

\SetKwComment{Comment}{$\triangleright$\ }{} 
\usepackage{authblk}
\usepackage{todonotes}
\presetkeys{todonotes}{inline}{}

\usepackage{array}
\newcolumntype{L}{>{\centering\arraybackslash}m{4cm}}
\newcolumntype{C}{>{\centering\arraybackslash}m{2cm}}

\icmltitlerunning{Generalized Reinforcement Meta Learning for Few-Shot Optimization}

\begin{document}

\twocolumn[
\icmltitle{Generalized Reinforcement Meta Learning for Few-Shot Optimization}



\icmlsetsymbol{equal}{*}

\begin{icmlauthorlist}
\icmlauthor{Raviteja Anantha}{aff1}
\icmlauthor{Stephen Pulman}{aff2}
\icmlauthor{Srinivas Chappidi}{aff1}
\end{icmlauthorlist}

\icmlaffiliation{aff1}{Apple Inc, Seattle, Washington, USA}
\icmlaffiliation{aff2}{Apple Inc, Cambridge, Cambridgeshire, UK}

\icmlcorrespondingauthor{Raviteja Anantha}{raviteja\_anantha@apple.com}

\icmlkeywords{Few-Shot Learning, Meta Learning, Reinforcement Learning}
\vskip 0.15in
]



\printAffiliationsAndNotice{}  

\begin{abstract}
We present a generic and flexible Reinforcement Learning (RL) based meta-learning framework for the problem of few-shot learning. 
During training, it learns the best optimization algorithm to produce a learner (ranker/classifier, etc) by exploiting stable patterns in loss surfaces. Our method implicitly estimates the gradients of a scaled loss function while retaining the general properties intact for parameter updates. Besides providing improved performance on few-shot tasks, our framework could be easily extended to do network architecture search. 
We further propose a novel dual encoder, affinity-score based decoder topology that achieves additional improvements to performance.
Experiments on an internal dataset, MQ2007, and AwA2 show our approach outperforms existing alternative approaches by 21\%, 8\%, and 4\% respectively on accuracy and NDCG metrics. On Mini-ImageNet dataset our approach achieves comparable results with Prototypical Networks. Empirical evaluations demonstrate that our approach provides a unified and effective framework. 
\end{abstract}

\section{Introduction}
The key idea of machine learning is to learn patterns from data, and it is important to use informative representations. One
of the areas where deep learning excels is through the ability to learn the representations automatically and
hierarchically. Structured representations remove the need of handcrafted features while capturing complex patterns from data. Typically the superior performance of deep learning requires massive amounts of training
data  and often performs poorly in small data settings. \textit{Few-shot learning} \cite{erik:2000,russ:2011,koch:15,Santoro:16,vinyals:16,ravi:17,finn:17,Jake:17} aims to produce an
effective model (ranker, classifier, etc) given only a few examples from each domain/class. After training, the model can generalize to new
domains/classes not seen in training. Although this is a hard task, humans are exceptionally good at it. This
learning technique also has practical applications. For example, in the context of Voice Assistants, often there is variance from Speech and NLU (Natural Language Understanding) sub-systems. A DM (Decision Making) sub-system consumes the outputs of Speech and NLU sub-systems to choose the best intent and DM should be able to preserve and leverage the knowledge from previous feature distributions. There are other practical applications like \textit{Robotics}~\cite{finnyu:17,gui:18} and \textit{Autonomous Systems}~\cite{xu:18} where learning capability of this sort is critical.

In the few-shot regime, the standard optimizers typically do not generalize well. In traditional model training,
typically one only focuses on the network blocks used in the models, \textit{viz.,} LSTMs, CNNs, activation functions,
tuning model parameters. But the optimization block mostly stays static: experts design it, iterating on
theoretical analysis and empirical validations. Some of the popular hand-engineered optimization algorithms are
AdaGrad~\cite{duchi:11}, RMSProp~\cite{tieleman:12}, and Adam~\cite{kingma:14}, to name a few. A step-size adaption scheme \cite{Michal:16} improves on these existing hand-engineered optimization algorithms. Attacking the few-shot problem by modifying optimization algorithms is a known approach. Optimization algorithm can be learned using guided policy search \cite{Ke:16}. Some approaches learn both the weight initialization and the optimizer \cite{ravi:17}, while few other approaches use a gradient obtained through a gradient to update parameters \cite{finn:17}.  

In this paper, we propose to learn the optimization algorithm by observing its execution and implicitly scaling the loss
surface while retaining the correct gradient direction intact to reach the best optimum and update model parameters accordingly.
We propose to learn an optimization algorithm that updates the model parameters in such a way that when executed on
test/blind data, it lands in the optimum on the respective loss-surface. In Section~3, we describe the meta-learning
framework to learn the optimization algorithm followed by introducing a novel encoder-decoder architecture in Section~4.

\section{Related Work}

Our work is an intersection of three critical areas of machine learning: \textit{meta-learning}, \textit{optimization} and \textit{few-shot learning}. Below we present related work in each of the three main fields of our work, and later we present related work that combines the three areas.

\textbf{Meta-Learning:}
\textit{Meta Learning} or \textit{Learning to Learn} \cite{thrun:95,thrun:98,vilalta:02,brazdil:08,thrun:12} has a long history, and it is an important building block in Artificial Intelligence. In this approach, we have a meta-learner and a learner (sub-system that learns from a meta-learner), where experience is gained by exploiting meta-knowledge extracted from a sequence of episodes on a dataset. There are two popular perspectives on how meta-knowledge should be learned, and we present them below.

One form is learning common patterns among a family of tasks presented in the training dataset so that the learner can quickly adapt to unseen tasks from the same family. This form of learning is also termed as \textit{Transfer Learning} or \textit{Multi-task Learning}. Another form is to learn the correlation between latent structures of tasks and different learners so that the meta-learner produces the best learner that achieves the best performance on the target task.

Networks that learn to modify their own weights over a number of update steps on an input have been studied well \cite{bengio:90, Schmidhuber:92, bengio:92, hochreiter:01, Andrychowicz:16}. The use of RL for meta-learning \cite{Duan:16, Wang:16}, and how network architecture search can be meta-learned using policy-gradient update \cite{zoph} have been explored. 

\textbf{Optimization:}
The move from manual feature engineering to an automated paradigm (deep learning) has been very successful. In spite of this, optimization algorithms are still designed by hand. The idea of automatic step-size
adaption for stochastic gradients is popular and different strategies were proposed; one line of work casts the learning rate as a parameter to train via gradient descent \cite{Baydin:17}, while another approach is to use $2^{nd}$ order information \cite{byrd:16, recht:17}. There have been meta-learning approaches to learn the optimization algorithm, where a gradient descent approach is used \cite{Andrychowicz:16}, and a guided policy search is used \cite{Ke:16}. Adding proposed parameters to the update rule \cite{Michal:16} improves over existing optimizers (like Adam \cite{kingma:14}).

\textbf{Few-Shot Learning:}
The best-performing methods for few-shot learning have been mainly metric learning methods. Siamese networks \cite{koch:15} train CNNs to encode data in such a way that data points from the same class are closer while from the
different classes are far apart. Matching networks \cite{vinyals:16} use recurrence with an attention
mechanism. Prototypical Networks \cite{Jake:17} compute a multi-dimensional representation through a learnable embedding
function which serves as a prototype for each class. Prototypical Networks are equivalent to Matching Networks for a
one-shot scenario while they differ in the few-shot case. 

\textbf{Optimization for Few-Shot Meta-Learning:}
The goal of few-shot meta learning is to train a learner (model) on a small dataset pertaining to some tasks, which can quickly adapt to new tasks. A memory-augmented neural network \cite{Santoro:16} can be trained to learn how to store and retrieve memories for each classification task. Approaches learning both weight initialization and the optimizer \cite{ravi:17}, for few-shot image recognition, has been shown to empirically outperform metric learning methods \cite{koch:15, vinyals:16}. In other approaches parameters are updated using a gradient through a gradient \cite{finn:17}.

Unlike these methods, our approach GRE-METL (Generalized REinforcement METa Learning) learns an optimization algorithm
based on policy gradients while also learning network architecture. Learning network architecture using RL is also
proposed in \cite{zoph}, we extended that work to learn optimization for the few-shot regime.

\section{Task Description and Methodology}
\label{sec:task-desc}
\subsection{Task Description}

Our goal is to generalize to different tasks (ranking, classification) coming from a distribution of tasks \textit{P(T)} that belongs to the
same family. In meta-learning, we work with meta-sets: we have a meta-set \textit{$D_{meta}$} which we assume is from
\textit{P(T)}, and we have \textit{$D_{meta-train}$} and \textit{$D_{test}$} from \textit{$D_{meta}$}. Traditionally in
few-shot tasks, we consider \textit{k-shot}, \textit{N-class} learning tasks, where for each \textit{$D_{train}$} (a
sample from \textit{$D_{meta-train}$}) we have \textit{k} labeled data points for each of \textit{N} classes. We want to
do well on \textit{$D_{test}$}. We first split the training data \textit{$D_{meta-train}$} into different
static mini-batches (\textit{$D_{train}$}) by random sampling. Dynamic mini-batches modify the intra-distribution of tasks during training in every meta-epoch which will make the training to not converge. Given a model with some parameters $f(X,\theta)$, each
task (\textit{$D_{train}$}) from a training set will have a \textit{loss surface} with respect to the \textit{parameters}
$\theta$. Similarly unseen tasks will also have their respective loss surfaces. The key idea is to find - instead of
optimal parameters for a specific \textit{train} loss surface - a way to get the optimal parameters that not only
perform better on seen loss surfaces but also on unseen loss surfaces pertaining to the same task family \textit{P(T)},
\textit{i.e.} \textit{L($f(X_{test};\theta)$, $Y_{test}$)} on the test loss surface should be close to the \textit{optimum}. Our
intuition here is that these loss surfaces would have common structures since they are from same task family, and basing
optimization on these commonalities should lead to better generalization. We empirically test our intuition on one internal and three benchmark datasets. 

\subsection{Methodology}
The core idea is to transform the loss surface \textit{$L$} to \textit{$L^{'}$} while retaining the general properties of the objective function. To illustrate, we take the simple example of a 2-dimensional convex function in Figure~\ref{fig:scaled_losses} where we obtain gradients of scaled loss surface to update parameters, this has two advantages:

\begin{itemize}
\item Captures commonalities among different loss surfaces from the same distribution of tasks to predict the best navigation path to the optimum.
\item Faster convergence.
\end{itemize}

To formulate, let $X,$ $Y,$ and $\theta$ denote feature vectors, labels, and model parameters respectively.

\begin{equation} \label{eq:1}
L = \textit{L($f(X;\theta)$, $Y$)}
\end{equation}

The transformer function $\phi$ scales the loss function $L$ to produce $L^{'}$. The 
transformed loss surface $L^{'}$ possess the same general properties as of $L$, and the gradients of $L^{'}$ build the path towards the optimum.

\begin{equation} \label{eq:2}
L^{'} = \phi(L;\lambda_1,..,\lambda_n)
\end{equation}

where $\lambda_{i}$ corresponds to a single or group of parameters, the latter is preferred when those parameters are
optimizing for the same \textit{sub-task}, and this is explained further in Section~5.

The gradients of the transformed loss function $L^{'}$ would then be used to update the model parameters:

\begin{equation} \label{eq:3}
\theta_{t} = \theta_{t-1} - \alpha \bigtriangledown_{\theta} L^{'}
\end{equation}

This algorithm is shown in Algorithm~\ref{alg:algorithm1}. But it is hard to formulate the transformer $\phi$. We show the empirical approximation for Algorithm \ref{alg:algorithm1} in Algorithm \ref{alg:algorithm2}. We propose an RL-based meta-learning scheme that learns an optimization algorithm based on a policy $\pi$ that implicitly does this transformation and obtains the gradients. The key insight here is since we are linearly transforming the loss function $L$ to get $L^{'}$, the gradient of $L^{'}$ is also some linear transformation of the gradient of $L$. Let the gradient-scaling coefficients be $\lambda^{'}_{i}$ corresponding to $\theta_{i}$, then the approximation of the gradients is given as:

\begin{figure*}
  \centering
  \begin{subfigure}[b]{0.3\linewidth}
    \includegraphics[width=\textwidth,trim=0.1mm 0.1mm 0.1mm 0.2mm, clip=true]{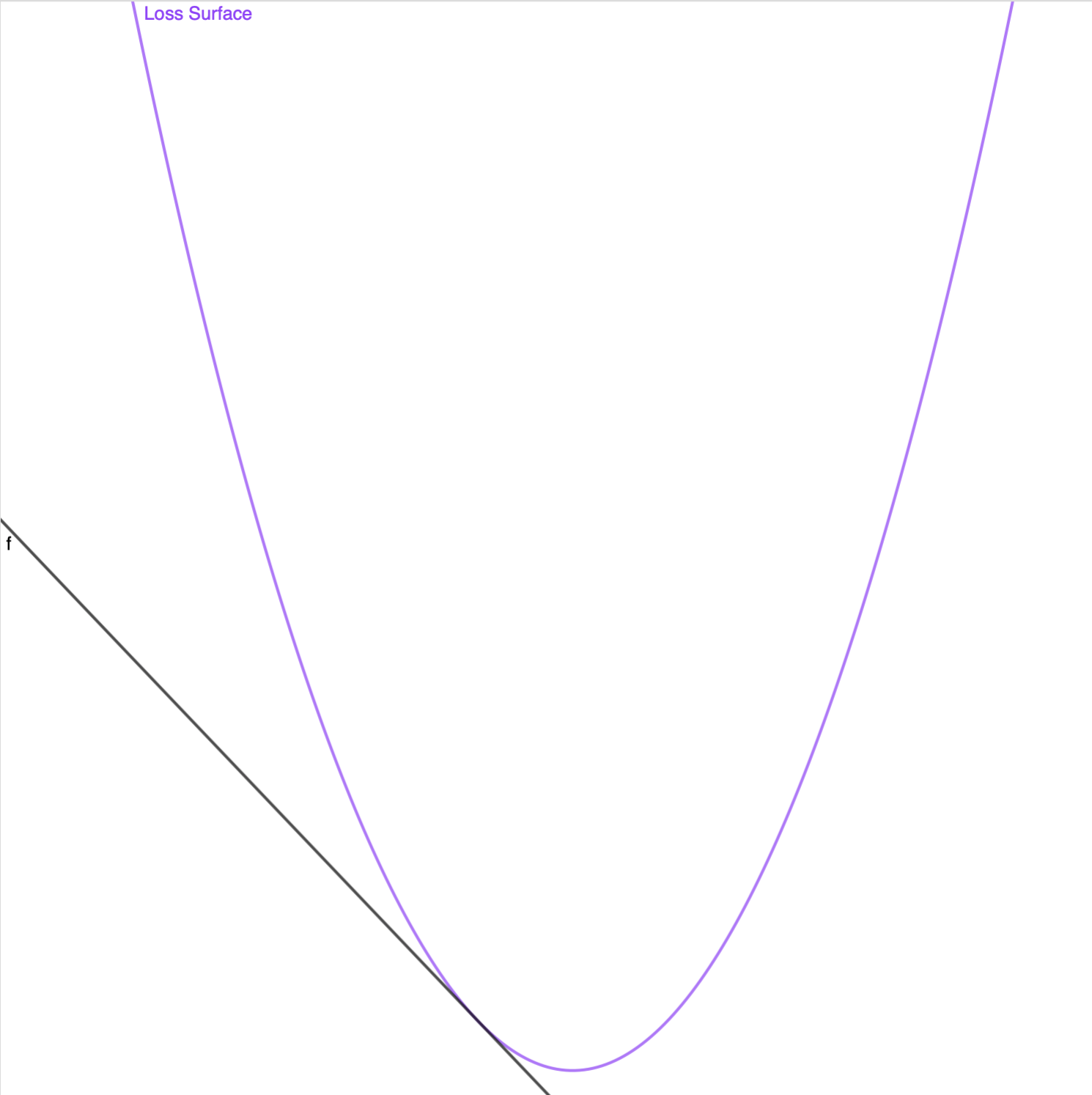}
    \caption{$L$}
  \end{subfigure}
  \begin{subfigure}[b]{0.3\linewidth}
    \includegraphics[width=\textwidth,trim=0.1mm 0.1mm 0.1mm 0.2mm, clip=true]{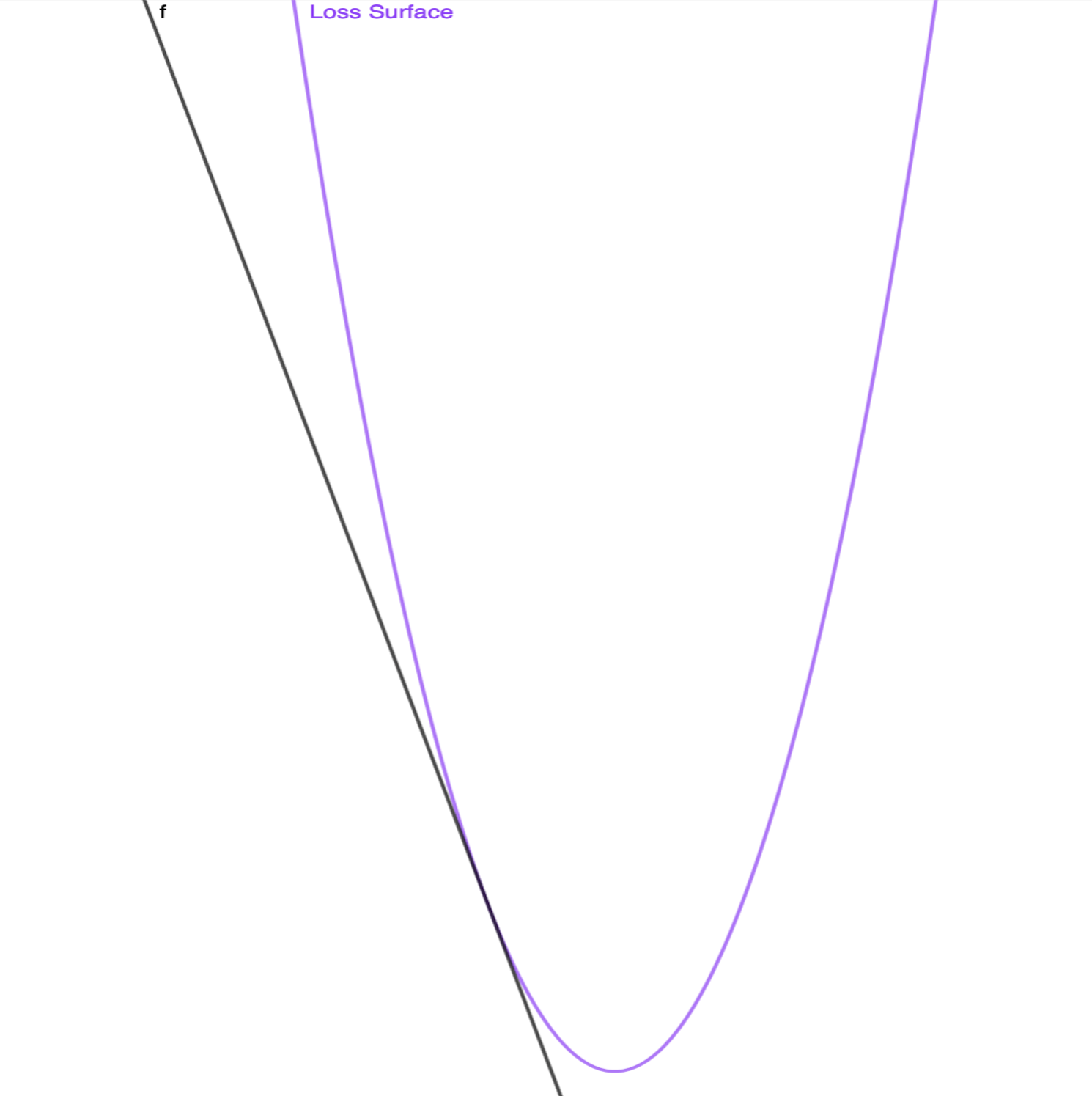}
    \caption{$L^{'}$ = $\phi(L;\lambda_1,..,\lambda_n)$}
  \end{subfigure}
  \caption{Actual and transformed loss surfaces with the same general properties.}
  \label{fig:scaled_losses}
\end{figure*}

\removelatexerror
\begin{algorithm}[H]
\SetAlgoLined

\textbf{Require:}~~$\textit{P(T)}$ \Comment*[r]{distribution of tasks}

\textbf{Require:}~~$\phi(.)$ \Comment*[r]{transformer function}

$\theta \gets \hspace{0.15cm} $random initialization;$ $

 \While{not done}{
  $ $sample$ \hspace{0.2cm} \textit{$D_{meta}$} \sim \textit{P(T)} ;$
  
  $ $get$ \hspace{0.2cm} \textit{$D_{meta-train}$} $, $ \textit{$D_{heldout}$} \hspace{0.2cm} $from$ \hspace{0.2cm} \textit{$D_{meta}$};$
  
  \ForAll{$\textit{$D_{train}$} \hspace{0.2cm} in \hspace{0.2cm} \textit{$D_{meta-train}$}$}{
  
  $Y^{'} \gets f(X;\theta)$ \Comment*[r]{predicted label}
  
  $L \gets L(Y^{'},Y)$ \Comment*[r]{compute loss}
  
  $\bigtriangledown_{\theta} L^{'}  \gets \bigtriangledown_{\theta} (\phi(L)) $ \Comment*[r]{compute gradients of transformed loss}
  
  $\theta_{t} \gets \theta_{t-1} - \alpha \bigtriangledown_{\theta} L^{'} $ \Comment*[r]{parameter update}
  
  $compute \hspace{0.2cm} and \hspace{0.2cm} track \hspace{0.2cm} \newline accuracy \hspace{0.2cm} for \hspace{0.2cm} (X_{heldout}, Y_{heldout}) \in D_{heldout} ; $
  
  }
  
 }
 \caption{Directly transforming the loss surface}
 \label{alg:algorithm1}
\end{algorithm}

\vskip 0.25in

\begin{equation} \label{eq:4}
\bigtriangledown^{'}_{\theta} L^{'} = \lambda^{'}_{i} \bigtriangledown_{\theta} L
\end{equation}

\begin{equation} \label{eq:5}
\theta_{t} = \theta_{t-1} - \alpha \bigtriangledown^{'}_{\theta} L^{'}
\end{equation}

\vskip 0.25in

We learn these gradient-scaling coefficients $\lambda^{'}_{i}$ through a meta-learner that is trained using RL. In every
\textit{meta-epoch} (epoch of meta-learner), it generates $\lambda^{'}_{i}$, and a sequence of model architecture parameters. This is done
based on the action picked by the  meta-learner from the space of actions \textit{A}, \textit{$a_{1}$} to
\textit{$a_{t}$}.  For every epoch of \textit{learner} (the sub-system that learns from the meta-learner) in a
meta-epoch, we compute the accuracy \textit{R} on the held-out dataset and use this as the reward signal for the
meta-learner. We use a policy-gradient method to iteratively update meta-learner parameters $\theta_{meta}$: the REINFORCE rule from \cite{Ronald:92},  also used in previous works \cite{zoph}, given below:

\begin{equation} \label{eq:6}
\bigtriangledown_{\theta_{meta}} P = \bigtriangledown_{\theta_{meta}} \log P(a_{t}|a_{(t-1):1};\theta_{meta}) R
\end{equation}

\begin{equation} \label{eq:7}
\bigtriangledown_{\theta_{meta}} L = \sum_{t=1}^{T} E_{P(A,\theta_{meta})} [\bigtriangledown_{\theta_{meta}} P]
\end{equation}

To minimize \textit{regret}, in every meta-epoch we randomly sample from a uniform distribution of [0.0, 1.0] and based on the threshold we either \textit{explore} or \textit{exploit}.

\removelatexerror
\begin{algorithm}[H]
\SetAlgoLined
\textbf{Require:}~~$\textit{$D_{meta}$} \hspace{0.2cm} $from$ \hspace{0.2cm} $\textit{P(T)}$ ;$ 

$ $get$ \hspace{0.2cm} \textit{$D_{meta-train}$} $, $\textit{$D_{heldout}$} $, $\textit{$D_{test}$} \hspace{0.2cm} $from$ \hspace{0.2cm} \textit{$D_{meta};$}$

$\theta_{meta} \gets \hspace{0.15cm} $random initialization$; $

define $\textit{$p_{explore}$}$ \Comment*[r]{probability to decide when to explore vs exploit}

 \While{meta-epochs not done}{
    
    explore $\gets$ random\_draw $\in$ [0.0, 1.0] \Comment*[r]{to minimize regret and trade-off exploration vs exploitation}
    
    \uIf{$(1^{st} \hspace{0.15cm} epoch) \hspace{0.15cm} \textbf{OR} \hspace{0.15cm} (explore\hspace{0.15cm}<\hspace{0.15cm}p_{explore})$ }
    {
        \Comment{explore}
        
        randomly choose $\lambda_{1}$, \dots, $\lambda_{n}$, and $\theta_{hyper};$
    
    }
    \uElse {
    
       \Comment{exploit}
    
       update policy $\pi$ with Equation-\ref{eq:7};
       
       meta-learner $M_{meta}$ picks actions $a_{\lambda}$ $\in$ $A_{\lambda}$ and $a_{hyper}$ $\in$ $A_{hyper}$ that generates $\lambda_{1}$ $\dots$ $\lambda_{n}$, and hyperparameters $\theta_{hyper}$ respectively;
       
    }
    
    \ForEach{$\textit{$D_{train}$} \hspace{0.2cm} $in$ \hspace{0.2cm} \textit{$D_{meta-train}$} $}{
    
    $Y^{'} \gets f(X;\theta)$ \Comment*[r]{predicted label}
    
    $L \gets L(Y^{'},Y)$ \Comment*[r]{compute loss}
    
    $ \bigtriangledown^{'}_{\theta} L^{'} =  \phi(\bigtriangledown_{\theta} L; \lambda^{'}_{1} \dots \lambda^{'}_{n} )$ \Comment*[r]{get scaled gradients}
    
    $ \theta_{t} = \theta_{t-1} - \alpha \bigtriangledown^{'}_{\theta} L^{'} $ \Comment*[r]{update learner's parameters with scaled gradients}
    
    $ R \gets getHeldOutAccuracy(.) $ 
    
    update policy $\pi$ using Equation-\ref{eq:7};
    
    meta-learner $M_{meta}$ picks action $a_{\lambda}$ $\in$ $A_{\lambda}$ that generates $\lambda_{1}$ $\dots$ $\lambda_{n}$;
    
    track accuracy on $ \textit{$D_{heldout}$} $;
    
    }
    
    $R \gets getTestAccuracy(.) $ \Comment*[r]{get reward on $\textit{$D_{test}$}$ from learner with best heldout accuracy}
 
 }

 \caption{Empirical approximation of Algorithm~1}
 \label{alg:algorithm2}
\end{algorithm}

\begin{figure*}
\centering
  \includegraphics[scale=0.35]{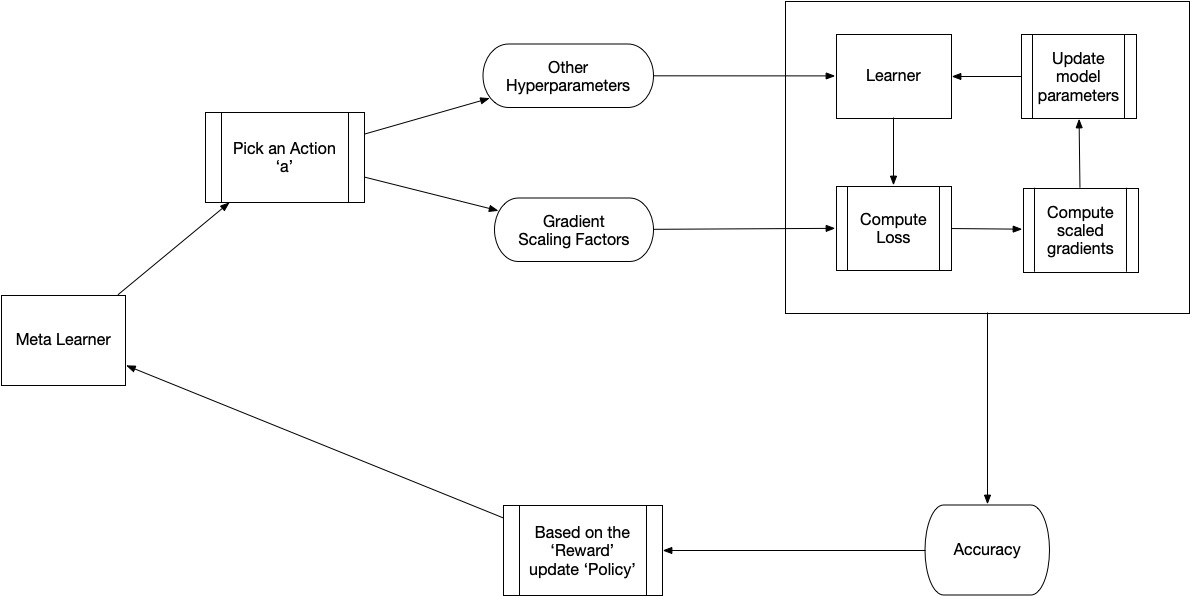}
  \caption{Flow Diagram of Algorithm~2. Meta Learner for every meta-epoch chooses from a set of actions that selects hyperparameters and gradient scaling factors for the Learner. Learner for every epoch updates its parameters using the loss scaled by gradient scaling factors. The best Learner model is chose based on a heldout dataset, and its accuracy is used as the Reward to update the Meta Learner's policy for choosing actions.}
  \label{fig:flow_diagram}
\end{figure*}

\section{Dual Encoder, Affinity-score based Decoder Topology}
\label{sec:enc-dec}
\subsection{Motivation}
In many real-world problems different groups of features capture various aspects of the input. For instance, in Voice
Assistants, \textit{semantic aspects} are extracted from ASR (Automatic Speech Recognition) and NLU modules. The outputs
of many of these components include score distributions and categorical values. \textit{Relevance} and
\textit{Executability} aspects are captured by \textit{user-data} and consists of categorical and numeric features. Conceptually learning the correlation among such feature groups
captured in a metric space is an efficient and effective way of decoding correct insights from unseen
domains/classes. We investigated the idea of hierarchically encoding these groups of features independently, and then
learning the correlation among them as captured by an \textit{affinity score vector} in metric space. Finally, we decode the \textit{affinity score vector} to predict the right action/class.

\subsection{Dual Encoders}
\begin{figure*}
\begin{center}
  \includegraphics[scale=0.4]{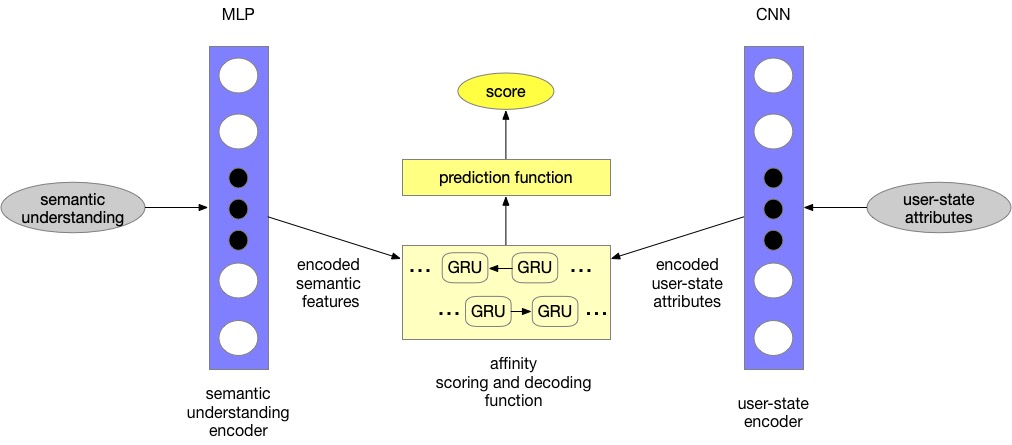}
  \caption{Dual Encoder, Affinity Decoder Topology used on the internal dataset. Input features are grouped into two categories: \textit{semantic understanding} and \textit{user-state} attributes. The former is fed into a Multi-Layer Perceptron (MLP) encoder, and the latter to a CNN encoder. The resulting encoded states are used in the Affinity Decoder where the decoded states are used in the prediction function to produce a score for the downstream tasks (classification, ranking, etc). For AwA2 dataset we use image features in the place of semantic features, and numeric attributes in the place of user-state attributes.}
  \label{fig:dual_enc_aff}
    \end{center}
\end{figure*}

We evaluated this architecture topology by training models on an internal dataset and Animals with Attributes 2 (AwA2) (see Table~\ref{table:int-results} and Table~\ref{table:awa-table} for results, and Section~\ref{sec:datasets} for datasets description). Since both these datasets have two feature groups we used two encoders. It
is trivial to expand this idea to multiple ($>2$) encoders. Embeddings are learned for different groups of features
from different encoders as shown in Figure \ref{fig:dual_enc_aff}. Different network architectures can be used for different encoders as best suited by the type of data. We observed that leaving the user-data features from the internal dataset and numeric attributes from AwA2 dataset as static (\textit{i.e.}, non-trainable) results in best performance.

\subsection{Affinity Decoder and Shared-trainable Initialization}
\begin{figure*}
\begin{center}
  \includegraphics[scale=0.4]{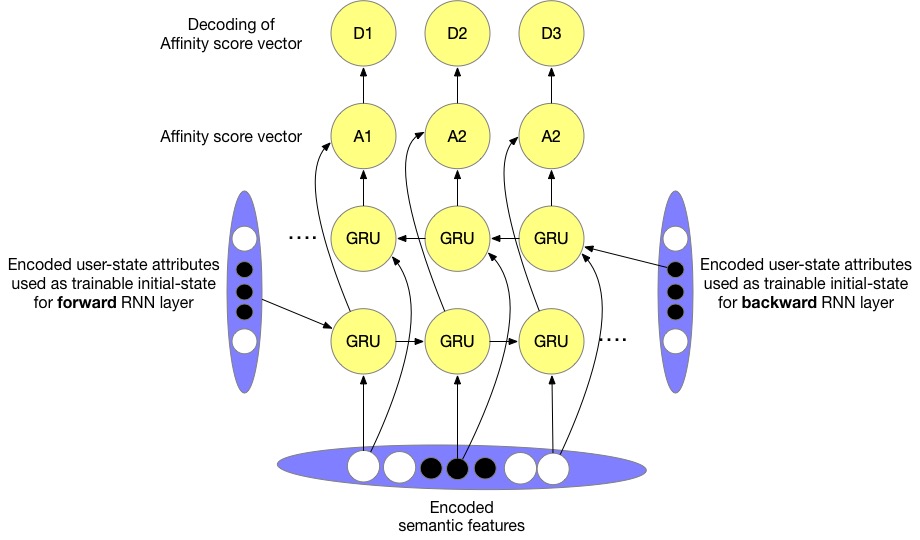}
  \caption{Affinity Decoder. Encodings for semantic features and user-state attributes are obtained using two different encoders. A bidirectional GRU layer is used in the affinity decoder. The initial hidden states of this layer are initialized using user-state attributes encoding, and the GRU cell unfolds on the semantic feature encoding sequence. The hidden states of the GRU layer are made trainable while restricting the update to only within the decoder and not propagating until the encoder's parameters. The affinity score vector (GRU layer output) is propagated to a linear layer to produce the decoder states.}
  \label{fig:aff_dec}
  \end{center}
\end{figure*}

Once we have the encodings for feature groups, we first learn the correlation among these encodings to learn the
affinity vector. We use a bidirectional RNN to learn the affinity-score function. We chose GRU \cite{cho:14} for RNN
cell since it has approximately the same capacity as of LSTM \cite{hochreiter:97} but with less number of parameters to
train. We observed that using a common attribute-value group as the initial state for the RNN cell resulted in better
performance. In our experiments, we had two such feature-group encodings, we used \textit{user-data} encoding for
internal dataset and \textit{numeric attribute values} encoding for \textit{AwA2} as the initial-state for the RNN layer
as shown in Figure-\ref{fig:aff_dec}. We make these initial states trainable but did not propagate back the updates to
the encoder parameters. This separation of encoder and decoder parameters improved the performance. This architecture is
robust to variance in distributional features since it normalizes by encoding and transforming the features to common
Hilbert space. We show the parameter updates in one GRU layer below, and the other layer follows similar updates. To
illustrate, let us consider the internal dataset. Let $x_{t}$ denote \textit{semantic-understanding} encoding and $h_{t-1}$ denote \textit{user-data} encoding. The \textit{gating scores} $G*_{t}$ are computed as follows with $W_{*}$ denoting corresponding weight matrices.

\begin{equation} \label{eq:8}
G1_{t} = \sigma ( W_{G1} \cdot [h_{t-1}, x_{t}] )
\end{equation}   

\begin{equation} \label{eq:9}
G2_{t} = \sigma ( W_{G2} \cdot [h_{t-1}, x_{t}] )
\end{equation}   

The candidate and actual hidden states are computed as:

\begin{equation} \label{eq:10}
\widetilde{h_{t}} = \tanh ( W \cdot [G2_{t} * h_{t-1}, x_{t}] )
\end{equation}   

\begin{equation} \label{eq:11}
h_{t} = (1 - G1_{t}) * h_{t-1} + G1_{t} * \widetilde{h_{t}}
\end{equation}   

The model trained with this architecture, hidden-state initialization and update procedure outperforms other
architectures both with/without meta-learning the optimizer in few-shot tasks as shown in Table~\ref{table:int-results} and Table~\ref{table:awa-table}.

\section{Experiments}
\subsection{Datasets}
\label{sec:datasets}
We empirically evaluate our approach on one internal dataset and three benchmark datasets for the task of few-shot learning.

\subsubsection{Internal Data}
Our internal dataset consists of 10,000 user requests to a voice assistant from six domains: music, videos, app-launch, and three knowledge-related domains. This dataset consists of all the features for intents from the suite of models from \textit{Speech Recognition} and \textit{Natural Language Understanding} (NLU) components. All data is anonymized and minimal information leaves the user's device.

Each request has a corresponding list of intents generated by the \textit{Speech} and \textit{NLU} components. Every intent has certain attribute values extracted from \textit{user-data}. We have 110 such attributes for these six domains of both categorical and numeric types. These attributes capture the \textit{relevance} and \textit{executability} aspects whereas features from \textit{Speech} and \textit{NLU} components capture the semantic-understanding aspects. Each intent for a user-request is assigned with a relevance grade internally by our annotators. 

We created 5 different combinations of the dataset where for each combination we had 3 training domains, 1 validation domain and 2 test domains. Every domain had 5-data points in training, 15 in both validation and test that are randomly sampled without replacement.

\subsubsection{Microsoft MQ2007}
The MQ2007 is one of the two query sets from LETOR (LEarning TO Rank) benchmark datasets.\footnote{\url{https://www.microsoft.com/en-us/research/project/letor-learning-rank-information-retrieval/}} The specific dataset we used is from LETOR 4.0 \cite{Tao:09} and consists of 1,700 queries with standard features, relevance labels. Since there is no concept of domains/classes in this dataset, we performed K-means clustering for \textit{K}=10, we chose \textit{K} by performing \textit{Silhouette analysis} with threshold for the \textit{measure} at 0.5. We created 5 different combinations of the dataset where each combination had 5 training domains, 2 validation domains and 3 test domains. Every domain had 5 data-points in training. Validation and test had 15 data-points that were randomly sampled without replacement.

\subsubsection{Mini-ImageNet}
The Mini-ImageNet dataset was originally proposed by \cite{vinyals:16}, since exact splits were not released then, \cite{ravi:17} introduced new procedure with 100 classes.\footnote{\url{http://www.image-net.org}} We follow their procedure of selecting random 100 classes: 64 training classes, 16 validation classes and 20 test classes to directly compare with their performance. We use 5 examples per class in training and 15 examples per class both in validation and test to perform 5-shot classification.

\subsubsection{Animals with Attributes 2 (AwA2)}
Animals with Attributes 2 \cite{XLSA18} is a widely-used benchmark dataset for \textit{transfer learning} algorithms.\footnote{\url{https://cvml.ist.ac.at/AwA2/}} It consists of 37,322 images with 50 classes and 85 numeric attribute values for each class. This dataset share similar characteristics with our internal dataset having separate attribute features. We created 5 different combinations of the dataset with each combination having 30 training classes, 5 validation classes and 15 test classes. We randomly sampled 5 examples from \textit{proposed split} \cite{XLSA18} for training and 15 examples for validation and testing each, per class.  

\subsection{Training Details}
For all the gradient-based methods we used the same hyperparameter tuning strategy which is a combination of grid and random search, and Adam optimizer with default parameters. We employed exponential decay for the learning-rate schedule. The exploration probability, $p_{explore}$, used in GRE-METL was grid-searched with a step-size of 0.1 in the range [0.0, 1.0]. The results were sensitive to the value of $p_{explore}$. On MQ2007 and Mini-ImageNet the optimal $p_{explore}$ values were 0.3 and 0.4 respectively, whereas on our internal dataset and AwA2 the optimal $p_{explore}$ value was 0.1.

For the internal and AwA2 datasets we trained a ranker and a classifier respectively using GRE-METL and encoder-decoder architecture as described in Algorithm \ref{alg:algorithm2}, Section~\ref{sec:task-desc} and Section~\ref{sec:enc-dec}. For training LambdaMART~\cite{burges:10} we used XGBoost implementation.\footnote{\url{https://github.com/dmlc/xgboost}} Meta-LSTM~\cite{ravi:17} was trained using the Torch implementation.\footnote{\url{https://github.com/twitter/meta-learning-lstm}} We used pairwise ranking loss for training ranker models and categorical cross-entropy for classifier models. MAML \cite{finn:17} was trained using the TensorFlow implementation.\footnote{\url{https://github.com/cbfinn/maml}} For training LambdaMART with MAML, we modified the parameter update to use lambda~\cite{burges:06} computed using the $1^{st}$ order derivative of the loss function instead of the originally proposed method of using gradient obtained through gradient.
Prototypical Networks were trained using PyTorch implementation.\footnote{\url{https://github.com/jakesnell/prototypical-networks}} 

Our encoder-decoder architecture was only used on our internal and AwA2 datasets since they share similar characteristics that can be leveraged in our proposed architecture, \textit{i.e.}, having different feature groups.

\section{Results}
Our primary results are summarized in Tables~\ref{table:int-results}, \ref{table:mq-table}, \ref{table:imagenet-table}, and \ref{table:awa-table}. $\pm$ denotes a 99\% confidence interval.
\subsection{Internal Data}
Although we have relevance grades for the intents and we use LTR (Learning to Rank) to choose the best intent, unlike in
traditional LTR problems, the top intent in the ranked intent list could be especially important, such as for some assistant contexts where giving only one result may be necessary. Hence we use the top intent accuracy as our metric. We show significant improvements over other techniques that are SOTA
(state-of-the-art) on comparable tasks both with and without meta-learning using our encoder-decoder
architecture. Table~\ref{table:int-results} shows the accuracy mean and 3-$\sigma$ (3 standard deviations) of the 5 different dataset combinations for 5-shot
training along with standard deviation. For brevity we refer to our encoder-decoder architecture as DEnc-ADec.

\begin{table}[h!]
	\caption{Evaluation results of the internal dataset with mean-accuracy and 99\% confidence intervals. All results in \textit{italics} use one or both of Dual-Encoder, Affinity Decoder architecture or/and GRE-METL algorithm.}
 	\resizebox{\columnwidth}{!}{
 	\begin{small}
	\label{table:int-results}
			\begin{tabular}{LL}
				\toprule
				Algorithm & Top Intent Accuracy (5-shot)\\
				\midrule
				Logistic Regression & 34.1\% $\pm$ 0.3\% \\
				Feed Forward 3-layer NN  & 34.3\% $\pm$ 0.7\% \\
				Hyp-Rank & 38.2\% $\pm$ 2.3\% \\
				LambdaMART & 43.7\% $\pm$ 0.1\%  \\
				DEnc-ADec & \textit{45.6\% $\pm$ 1.2\% }  \\
				Meta-LSTM w/ DEnc-ADec & \textit{54.2\% $\pm$ 0.6\%} \\
				GRE-METL w/ LambdaMART & \textit{57.3\% $\pm$ 1.4\% } \\
				MAML with DEnc-ADec & \textit{61.4\% $\pm$ 0.9\%} \\
				GRE-METL w/ DEnc-ADec &  \textbf{\textit{65.1\% $\pm$ 1.5\%}} \\
				\bottomrule
			\end{tabular}
 	\end{small}
     \vspace{2mm}
     }
\end{table}

\subsection{Microsoft MQ2007}
On MQ2007 dataset we trained ranking models on the 5 different dataset combinations for 10-way, 5-shot ranking task. We use NDCG (Normalized Discounted Cumulative Gain) as our metric and we show significant improvements over SOTA approaches on both NDCG@1 and NDCG@5, shown in Table~\ref{table:mq-table}. 

\begin{table}[h!]
	\caption{Evaluation results of the MQ2007 dataset.}
 	\resizebox{\columnwidth}{!}{
 	\begin{small}
	\label{table:mq-table}
			\begin{tabular}{LCC}
				\toprule
				Algorithm & Test NDCG@1 (5-shot) & Test NDCG@5 (5-shot) \\
				\midrule
				LambdaMART & 67\% $\pm$ 0.2\% & 71\% $\pm$ 0.1\% \\
				Meta-LSTM w/ LambdaMART & 52\% $\pm$ 4.5\% & 56\% $\pm$ 1.8\% \\
				MAML w/ LambdaMART & 54\% $\pm$ 3.2\% & 73\% $\pm$ 1.7\% \\
				GRE-METL w/ LambdaMART & \textit{\textbf{73\% $\pm$ 2\%}} & \textit{\textbf{81\% $\pm$ 2.5\%}} \\
				\bottomrule
			\end{tabular}
 	\end{small}
     \vspace{2mm}
     }
\end{table}

\subsection{Mini-ImageNet}
We trained ResNet classifier as our baseline and improved it with other meta-learning approaches including ours, we also trained Prototypical Networks \cite{Jake:17}. We report 5-shot classification accuracies on this dataset, shown in Table~\ref{table:imagenet-table}. We significantly outperform all existing approaches, and perform slightly better than Prototypical Networks.

\begin{table}[h!]
	\caption{Evaluation results of the Mini-ImageNet dataset.}
 	\resizebox{\columnwidth}{!}{
 	\begin{small}
	\label{table:imagenet-table}
			\begin{tabular}{LL}
				\toprule
				Algorithm &  Classification Accuracy (5-shot) \\
				\midrule
				Baseline ResNet & 52\% $\pm$ 1.4\% \\
				Meta-LSTM w/ ResNet & 61.2\% $\pm$ 0.8\% \\
				MAML w/ ResNet & 63.1\% $\pm$ 0.4\% \\
				Prototypical Networks & \textbf{69.4\% $\pm$ 0.8\%} \\
				GRE-METL w/ ResNet & \textit{\textbf{71.10\% $\pm$ 1.7\%}}\\
				\bottomrule
			\end{tabular}
 	\end{small}
     \vspace{2mm}
     }
\end{table}

\subsection{Animals with Attributes 2 (AwA2)}
For AwA2 dataset we trained baseline ResNet and improved it with meta-learning algorithms. We also trained Prototypical Networks. Our encoder-decoder architecture combined with meta-learning specifically with GRE-METL shows significant improvements in Table~\ref{table:awa-table} on 5-shot accuracies.

\begin{table}[h!]
	\caption{Evaluation results of the AwA2 dataset.}
 	\resizebox{\columnwidth}{!}{
 	\begin{small}
	\label{table:awa-table}
			\begin{tabular}{LL}
				\toprule
				Algorithm & Classification Accuracy (5-shot) \\
				\midrule
				Baseline ResNet & 37.4\% $\pm$ 2.7\% \\
				Meta-LSTM w/ ResNet & 47.8\% $\pm$ 0.8\% \\
				MAML w/ ResNet & 57.2\% $\pm$ 1.3\% \\
				Prototypical Networks & 66.4\% $\pm$ 0.6\% \\
				GRE-METL w/ ResNet & \textit{67.5\% $\pm$ 0.5\%} \\
				GRE-METL w/ DEnc-ADec & \textit{\textbf{70.2\% $\pm$ 1.3\%}} \\
				\bottomrule
			\end{tabular}
 	\end{small}
     \vspace{2mm}
     }
\end{table}

\section{Conclusion}
We introduced a novel, generic, and flexible meta-learning scheme for few-shot learning based on the idea that we can learn the optimization algorithm by implicitly transforming the loss surface while retaining the general properties. We also showed how the framework is easily extended to perform network architecture search. We empirically show our approach sets new state-of-the-art on AwA2 and MQ2007 datasets and achieves comparable results with Prototypical Networks on Mini-ImageNet.

\section{Future Work}
While the current approaches including the proposed method achieve good performance on the desired evaluation metrics, it is not clear how to add a \textit{compactness} metric as a reward along with the \textit{desired evaluation} metric. To have a more practical reach and impact of these models it is important to deploy them on-device. This also helps preserve user-privacy and improve latency by saving round-trips to the server. Hence it is necessary to build compact models that can run on the device, especially the ones with limited compute resources.  We plan to investigate how to achieve the desired compactness characteristic of the model with minimal or no drop in the model performance.

\section*{Acknowledgements}
We would like to thank our colleagues Russ Webb, Dennis DeCoste, Gary Patterson, and Arturo Argueta for their insightful comments.


\bibliography{gre-metl}
\bibliographystyle{icml2020}

\end{document}